\documentclass[letterpaper]{article}
\usepackage{uai2018}
\usepackage[margin=1in]{geometry}

\usepackage{times}

\usepackage[dvips]{graphicx}
\usepackage{amssymb,amsmath,color}
\usepackage{url}
\usepackage{comment}

\usepackage[mathcal]{eucal}
\usepackage{multicol}
\usepackage[round]{natbib}
\usepackage{hyperref}       

\usepackage{color}
\definecolor{brightmaroon}{rgb}{0.76, 0.13, 0.28}
\definecolor{darkblue}{rgb}{0,0.08,0.45}
\hypersetup{
    colorlinks=true, 
    linkcolor=brightmaroon,  
    citecolor=darkblue,  
    filecolor=brightmaroon,  
    urlcolor=brightmaroon,   
}

\newcommand{\BEAS}{\begin{eqnarray*}}
\newcommand{\EEAS}{\end{eqnarray*}}
\newcommand{\BEA}{\begin{eqnarray}}
\newcommand{\EEA}{\end{eqnarray}}
\newcommand{\BEQ}{\begin{equation}}
\newcommand{\EEQ}{\end{equation}}
\newcommand{\BIT}{\begin{itemize}}
\newcommand{\EIT}{\end{itemize}}
\newcommand{\BNUM}{\begin{enumerate}}
\newcommand{\ENUM}{\end{enumerate}}
\newcommand{\BA}{\begin{array}}
\newcommand{\EA}{\end{array}}

\newcommand{\rb}{\mathbb{R}}

\newcommand{\mysec}[1]{Section~\ref{sec:#1}}
\newcommand{\eq}[1]{Eq.~(\ref{eq:#1})}
\newcommand{\myfig}[1]{Figure~\ref{fig:#1}}

\newtheorem{proposition}{Proposition}

\def \E{{\mathbb E}}
\def \P{{\mathbb P}}

\def \X{{\mathcal X}}

\def \R{{\mathbb R}}

\def \FF{{\mathcal F}}

\usepackage{times}

\title{Constant Step Size Stochastic Gradient Descent \\[.1cm] for Probabilistic Modeling}

\author{ {\bf Dmitry Babichev } \\
INRIA - ENS \\
PSL Research University\\
Paris, France \\
dmitry.babichev@inria.fr
\And
{\bf Francis Bach}   \\
INRIA - ENS \\
PSL Research University\\
Paris, France \\
francis.bach@inria.fr
}

\begin{document}

\maketitle

\begin{abstract}
Stochastic gradient methods enable learning probabilistic models from large amounts of data. While large step-sizes (learning rates) have shown to be best for least-squares  (e.g., Gaussian noise) once combined with parameter averaging, these are not leading to convergent algorithms in general. In this paper, we consider generalized linear models, that is, conditional models based on exponential families. We propose   averaging moment parameters instead of natural parameters for constant-step-size stochastic gradient descent.
 For finite-dimensional models, we show that this can sometimes (and surprisingly) lead to better predictions than the best linear model.
  For infinite-dimensional models, we show  that it always converges to optimal predictions, while averaging natural parameters never does.
  We illustrate our findings  with simulations on synthetic data and classical benchmarks with many observations.

\end{abstract}

\section{INTRODUCTION}

Faced with large amounts of data, efficient parameter estimation remains one of the key bottlenecks in the application of probabilistic models. Once cast as an optimization problem, for example through the maximum likelihood principle, difficulties may arise from the size of the model, the number of observations, or the potential non-convexity of the objective functions, and often all three~\citep{daphne_koller_book,kevin_murphys_book}.

In this paper we focus primarily on situations where the number of observations is large; in this context, stochastic gradient descent (SGD) methods which look at one sample at a time are usually favored for their cheap iteration cost. However, finding the correct step-size (sometimes referred to as the learning rate) remains a practical and theoretical challenge, for probabilistic modeling but also in most other situations beyond maximum likelihood~\citep{bottou2016optimization}.

In order to preserve convergence, the step size $\gamma_n$ at the $n$-th iteration typically has to decay with the number of gradient steps (here equal to the number of data points which are processed), typically as $C / n^\alpha$ for $\alpha \in [1/2,1]$ \citep[see, e.g.,][]{gradsto,bottou2016optimization}. However, these often leads to slow convergence and the choice of $\alpha$ and $C$ is difficult in practice. More recently, constant step-sizes have been advocated for their fast convergence towards a neighborhood of the optimal solution~\citep{Bac_Mou_2013}, while it is a standard practice in many areas~\citep{deep_learning_book_bengio}. However, it is not convergent in general and thus small step-sizes are still needed to converge to a decent estimator.

Constant step-sizes can however be made to converge in one situation. When the functions to optimize are quadratic, like for least-squares regression, using a constant step-size combined with an averaging of all estimators along the algorithm can be shown to converge to the global solution with the optimal convergence rates~\citep{Bac_Mou_2013,Die_Bac_2015}.

The goal of this paper is to explore the possibility of such global convergence  with a constant step-size in the context of probabilistic modeling with exponential families, e.g., for logistic regression or Poisson regression~\citep{mccullagh1984generalized}. This would lead to  the possibility of using probabilistic models (thus with a principled quantification of uncertainty) with rapidly converging algorithms. Our main novel idea is to replace the averaging of the \emph{natural} parameters of the exponential family by the averaging of the \emph{moment} parameters, which can also be formulated as averaging \emph{predictions} instead of \emph{estimators}. For example, in the context of predicting binary outcomes in $\{0,1\}$ through a Bernoulli distribution, the moment parameter is the probability $p \in [0,1]$ that the variable is equal to one, while the natural parameter is the ``log odds ratio'' $\log \frac{p}{1-p}$, which is unconstrained. This lack of constraint is often seen as a benefit for optimization; it turns out that for stochastic gradient methods, the moment parameter is better suited to averaging. Note that for least-squares, which corresponds to modeling with the Gaussian distribution with fixed variance, moment and natural parameters are equal, so it does not make a difference.

More precisely, our main contributions are:
\BIT
\item  For generalized linear models, we propose in \mysec{cond} averaging moment parameters instead of natural parameters for constant-step-size stochastic gradient descent.

\item For finite-dimensional models, we show in \mysec{finited} that this can sometimes (and surprisingly) lead to better predictions than the best linear model.
\item For infinite-dimensional models, we show in \mysec{kernels} that it always converges to optimal predictions, while averaging natural parameters never does.
\item We illustrate our findings in \mysec{exp} with simulations on synthetic data and classical benchmarks with many observations.
\EIT

\section{CONSTANT STEP SIZE STOCHASTIC GRADIENT DESCENT}
\label{sec:sgd}

\begin{figure}
\includegraphics[width=0.5\textwidth]{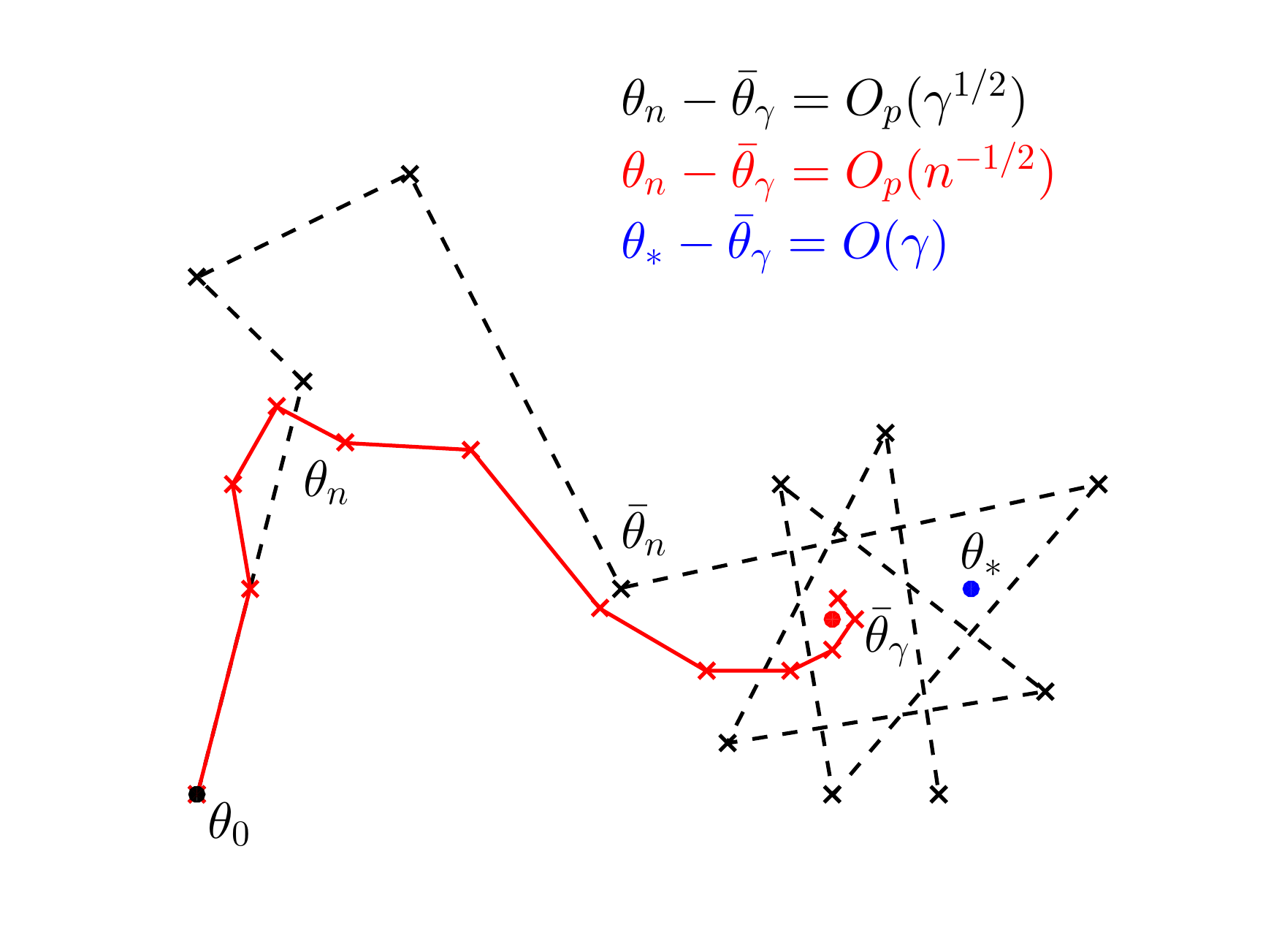}

\vspace*{-1cm}
\caption{Convergence of iterates $\theta_{n}$ and averaged iterates $\bar{\theta}_{n}$ to the mean $\bar{\theta}_\gamma$ under the stationary distribution~$\pi_\gamma$.}
\label{fig:mc}
\end{figure}

In this section, we present the main intuitions behind stochastic gradient descent (SGD) with constant step-size. For more details, see~\citet{dieuleveut2017bridging}.
We consider a real-valued function $F$ defined on the Euclidean space $\rb^d$ (this can be generalized to any Hilbert space, as done in \mysec{kernels} when considering Gaussian processes and positive-definite kernels), and a sequence of random functions $(f_n)_{n \geqslant 1}$ which are independent and identically distributed and such that $\E f_n (\theta) = F(\theta)$ for all $\theta \in \rb^d$. Typically, $F$ will the expected negative log-likelihood on unseen data, while $f_n$ will be the negative log-likelihood for a single observation. Since we require \emph{independent} random functions, we assume that we make single pass over the data, and thus the number of iterations is equal to the number of observations.

Starting from an initial $\theta_0 \in \rb^d$, then SGD will perform the following recursion, from $n=1$ to the total number of observations:
\BEQ
\label{eq:sgd} \theta_n = \theta_{n-1} - \gamma_n \nabla f_n(\theta_{n-1}).
\EEQ
Since the functions $f_n$ are independent, the iterates $(\theta_n)_n$ form a Markov chain. When the step-size $\gamma_n$ is constant (equal to $\gamma$) and the functions $f_n$ are identically distributed, the Markov chain is \emph{homogeneous}. Thus, under additional assumptions \citep[see, e.g.,][]{dieuleveut2017bridging,Mey_Twe_1993}, it converges in distribution to a stationary distribution, which we refer to as $\pi_\gamma$. These additional assumptions include that $\gamma$ is not too large (otherwise the algorithm diverges) and in the traditional analysis of step-sizes for gradient descent techniques, we analyze the situation of small $\gamma$'s (and thus perform asymptotic expansions around $\gamma=0$).

The distribution $\pi_\gamma$ is in general not equal to a Dirac mass, and thus, constant-step-size SGD is \emph{not} convergent.
However,  averaging along the path of the Markov chain has some interesting properties. Indeed, several versions of the ``ergodic theorem'' \citep[see, e.g.,][]{Mey_Twe_1993} show that for functions $g$ from $\rb^d$ to any vector space, then the empirical average  $\frac{1}{n} \sum_{i=1}^n g(\theta_i)$ converges in probability to the expectation $\int g(\theta) d\pi_\gamma(\theta)$ of $g$ under the stationary distribution $\pi_\gamma$. This convergence can also be quantified by a central limit theorem with an error whichs tends to a normal distribution with variance proportional equal to a constant times $1/n$.

Thus, if denote $\bar{\theta}_n = \frac{1}{n+1} \sum_{i=0}^n \theta_i$, applying the previous result to the identity function $g $, we immediately obtain that $\bar{\theta}_n$ converges to $\bar{\theta}_\gamma = \int \theta d\pi_\gamma(\theta)$, with a squared error converging in $O(1/n)$. The key question is the relationship between $\bar{\theta}_\gamma $ and the global optimizer~$\theta_\ast$ of $F$, as this characterizes the performance of the algorithm with an infinite number of observations.

By taking expectations in \eq{sgd}, and taking a limit with~$n$ tending to infinity we obtain that
\BEQ
\label{eq:stat} \int \nabla F (\theta) d\pi_\gamma(\theta) = 0,
\EEQ
that is, under the stationary distribution $\pi_\gamma$, the average gradient is zero. When the gradient is a linear function (like for a quadratic objective $F$), this leads to $ \nabla F ( \int\theta d\pi_\gamma(\theta) ) = \nabla F(\bar{\theta}_\gamma)= 0$, and thus $\bar{\theta}_\gamma$ is a stationary point of $F$ (and hence the global minimizer if $F$ is convex). However this is not true in general.

As shown by \citet{dieuleveut2017bridging}, the deviation $\bar{\theta}_\gamma - \theta_\ast$ is of order $\gamma$, which is an improvement on the non-averaged recursion, which is at average distance $O(\gamma^{1/2})$ (see an illustration in \myfig{mc}); thus, small or decaying step-sizes are needed. In this paper, we explore alternatives which are not averaging the estimators $\theta_1,\dots,\theta_n$, and rely instead on the specific structure of our cost functions, namely negative log-likelihoods.

\section{WARM-UP: EXPONENTIAL FAMILIES}
\label{sec:uncond}
In order to highlight the benefits of averaging moment parameters, we first consider unconditional exponential families. We thus consider the standard exponential family 
$q(x|\theta) = h(x) \exp ( \theta^\top T(x) - A(\theta) )$, where $h(x)$ is the base measure, $T(x) \in \rb^d$ is the sufficient statistics and $A$ the log-partition function. The function $A$ is always convex \citep[see, e.g.,][]{daphne_koller_book,kevin_murphys_book}. Note that we do not assume that the data distribution $p(x)$ comes from this exponential family. The expected (with respect to the input distribution $p(x)$) negative log-likelihood is equal to
\BEAS
F(\theta) & = &  - \E_{p(x)} \log q(x|\theta) \\
& =  &
A(\theta) - \theta^\top \E_{p(x)} T(x) - \E_{p(x)}\log h(x).
\EEAS
It is known to be minimized by $\theta_\ast $ such that $\nabla A(\theta_\ast) =  \E_{p(x)} T(x) $. Given i.i.d.~data $(x_n)_{n \geqslant 1}$ sampled from $p(x)$, then
the SGD recursion from \eq{sgd} becomes:
$$
\theta_n  = \theta_{n-1} - \gamma  \big[ \nabla A(\theta_{n-1}) - T(x_n) \big],
$$
while the stationarity equation in \eq{stat} becomes
$$ \int \big[ \nabla A(\theta) -    \E _{p(x)}T(x)] d\pi_\gamma(\theta)  = 0,$$
which leads to
$$ \int  \nabla A(\theta) d\pi_\gamma(\theta)  =  \E_{p(x)}T(x) = \nabla A(\theta_\ast).$$

Thus, averaging $\nabla A(\theta_n)$ will converge to $\nabla A(\theta_\ast)$, while averaging $\theta_n$ will \emph{not} converge to $
\theta_\ast$. This simple observation is the basis of our work.

Note that in this context of unconditional models, a simpler estimator exists, that is, we can simply compute the empirical average $\frac{1}{n} \sum_{i=1}^n T(x_i)$ that will converge to $\nabla A(\theta_\ast)$. Nevertheless, this shows that averaging moment parameters $\nabla A(\theta)$ rather than natural parameters $\theta$ can bring convergence benefits. We now turn to conditional models, for which no closed-form solutions exist.

\section{CONDITIONAL EXPONENTIAL FAMILIES}

\label{sec:cond}

Now we consider the conditional exponential family $q(y|x,\theta)=h(y)\exp\big(y\cdot \eta_\theta(x) - a(\eta_\theta(x))\big)$. For simplicity we consider only one-dimensional families where $y \in \rb$---but our framework would also extend to more complex models such as conditional random fields~\citep{lafferty2001conditional}. We will also assume that $h(y) = 1$ for all $y$ to avoid carrying constant terms in log-likelihoods. We consider  functions of the form $\eta_\theta(x)=\theta^\top\Phi(x)$, which are linear in a feature vector $\Phi(x)$, where $\Phi: \mathcal{X} \to \rb^d$ can be defined on an arbitrary input set $\mathcal{X}$. Calculating the negative log-likelihood, one obtains:
$$
f_n(\theta)    =    - \log q(y_n | x_n \theta)  =  -y_n\Phi(x_n)^\top\theta + a\big(\Phi(x_n)\theta\big) ,
$$
and, for any distribution $p(x,y)$, for which $p(y|x)$   may not be a member of the conditional exponential family,
\BEAS
F(\theta) & = &  \E_{p(x_n,y_n)} f_n(\theta) \\
& = &  \E_{p(x_n,y_n)}\Big[ -y_n\Phi(x_n)^\top\theta + a\big(\Phi(x_n)\theta\big)  \Big]. 
\EEAS
The goal of estimation in such generalized linear models is to find an unknown parameter $\theta$   given $n$ observations $(x_i,y_i)_{i=1,\dots,n}$:
\begin{equation}
\theta_{\ast} = \arg\min\limits_{\theta\in\rb^d}F(\theta).
\label{thetastar}
\end{equation}

\subsection{FROM ESTIMATORS TO PREDICTION FUNCTIONS}

Another point of view is to consider that an estimator $\theta \in \rb^d$ in fact defines a  function 
$\eta: \mathcal{X} \to \rb$, with value a natural parameter for the exponential family $q(y) =    \exp( \eta y - a(\eta))$.
This particular choice of function $\eta_\theta$ is linear in $\Phi(x)$, and we have, by decomposing the joint probability $p(x_n,y_n)$ in two (and dropping the dependence on $n$ since we have assumed i.i.d.~data):
\BEAS
  F(\theta)  \!\!\!  & = &  \!\!\! \E_{p(x)}\Big(  \E_{p(y | x) } \big[ 
 - y \Phi(x)^\top \theta + a ( \Phi(x)^\top \theta )    \big]
\Big) \\
\!\!\! & = &\!\!\!   \E_{p(x)} \Big(   - \E_{p(y | x) } y   \Phi(x)^\top \theta + a ( \Phi(x)^\top \theta ) 
\Big) \\
\!\!\! & = &\!\!\!  \mathcal{F} ( \eta_\theta), 
\EEAS
with $\mathcal{F}(\eta) = \E_{p(x)} \big(   -  \E_{p(y| x) } y \cdot \eta(x) + a ( \eta(x) ) 
\big) $ is the performance measure defined for   a \emph{function} $\eta: \X\to \rb$. By definition
$F(\theta) = \mathcal{F} ( \eta_\theta) = \mathcal{F}( \theta^\top \Phi(\cdot) )$.

However, the global minimizer of  $\mathcal{F} ( \eta)$ over all functions $\eta: \X \to \rb$ may not be attained at a linear function in~$\Phi(x)$ (this can only be the case if the linear model is well-specified or if the feature vector $\Phi(x)$ is flexible enough).
Indeed, the global minimizer of $\mathcal{F}$ is the function $\eta_{\ast\ast}: x \mapsto (a')^{-1}(  \E_{p(y | x) } y  )$ (starting from $\mathcal{F}(\eta) =  \int\big[a(\eta(x)) - \mathbb{E}_{p(x|y)}y\cdot \eta(x) \big] p(x)dx$ and writing down the Euler - Lagrange equation: $\frac{\partial \mathcal{F}}{\partial \eta} - \frac{d}{dx}\frac{\partial F}{\partial \eta'} = 0 \Leftrightarrow \big[a'(\eta) - \mathbb{E}_{p(x|y)}y\big] p(x) =0 $ and finally $\eta \mapsto (a')^{-1}(\mathbb{E}_{p(x|y)}y)$) and is typically not a linear function in $\Phi(x)$ (note here that  $p(y|x)$ is the conditional data-generating distribution).

\begin{figure}
\includegraphics[width=0.5\textwidth]{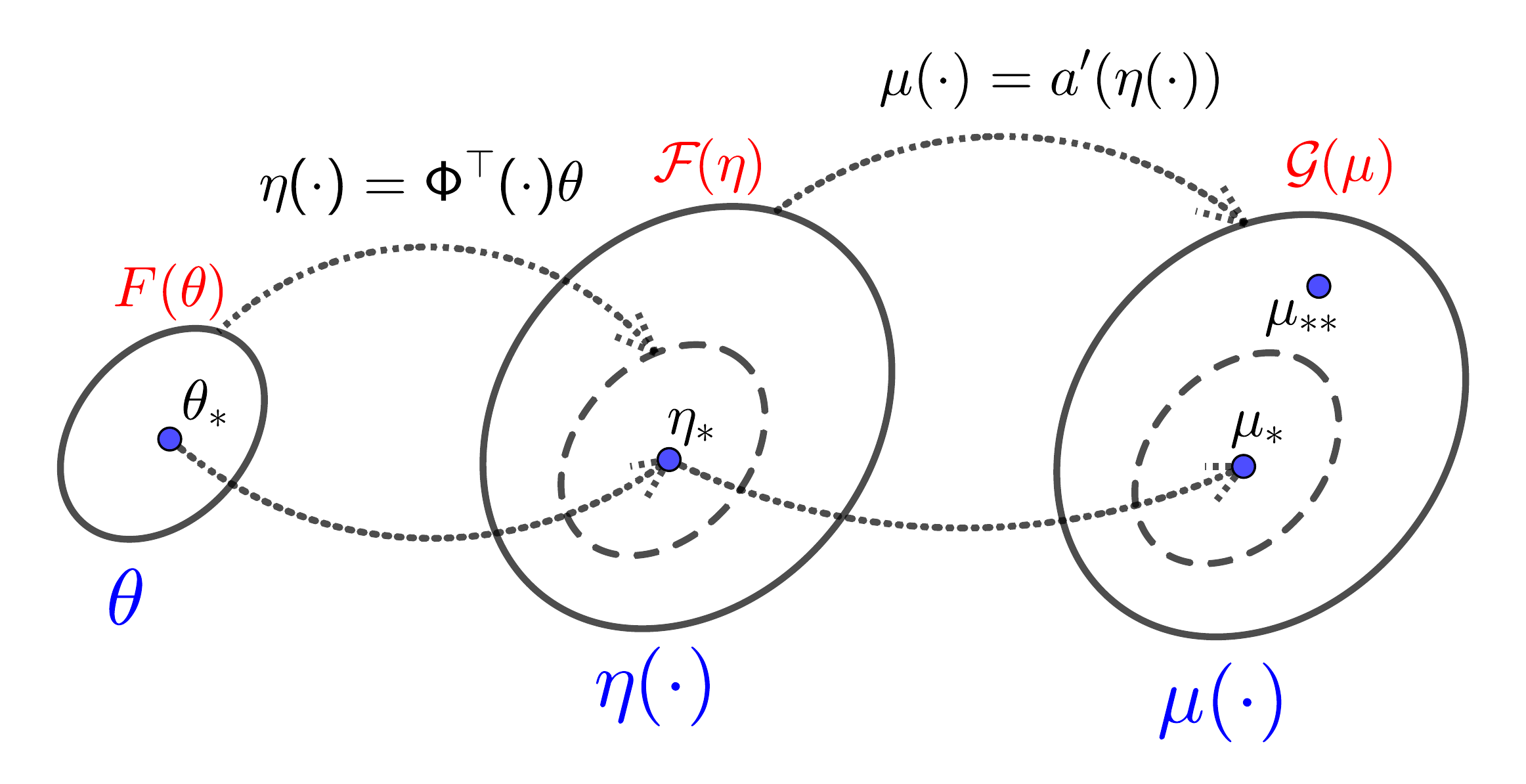}

\vspace*{-.415cm}

\caption{Graphical representation of reparametrization: firstly we expand the class of functions, replacing parameter $\theta$ with function $\eta(\cdot) = \Phi^\top(\cdot)\theta$ and then we do one more reparametrization: $\mu(\cdot) = a'(\eta(\cdot))$. Best linear prediction $\mu_{\ast}$ is constructed using $\theta_\ast$ and the global minimizer of $\mathcal{G}$ is $\mu_{\ast\ast}$. Model is well-specified if and only if $\mu_\ast = \mu_{\ast\ast}$. }
\label{Graph010}
\end{figure}

The function $\eta$ corresponds to the \emph{natural} parameter of the exponential family, and it is often more intuitive to consider  the \emph{moment} parameter, that is defining functions $\mu: \mathcal{X} \to \rb$ that now correspond to moments of outputs~$y$; we will refer to them as \emph{prediction functions}.
Going from natural to moment parameter is known to be done through the gradient of the log-partition function, and we thus consider for $\eta$ a function from $\mathcal{X}$ to $\rb$, $\mu(\cdot) = a'(\eta(\cdot))$, and this leads to  the performance measure
$$\mathcal{G}(\mu) = \mathcal{F}( (a')^{-1}(\mu(\cdot))).$$

Note now, that the global minimum of $\mathcal{G}$ is reached at $$\mu_{\ast\ast}(x) = \E_{p(y|x)} y.$$
We introduce also the prediction function $\mu_{\ast}(x)$ corresponding to the best $\eta$ which is linear in $\Phi(x)$, that is:
$$\mu_\ast(x) = a'\big(\theta_\ast^\top \Phi(x)\big).$$

We say that the model is well-specified when $\mu_\ast = \mu_{\ast \ast}$, and for these models,
$\inf_{\theta} F(\theta) = \inf_{\mu} \mathcal{G}(\mu)$. However, in general, we only have $\inf_{\theta} F(\theta) > \inf_{\mu} \mathcal{G}(\mu)$ and (very often) the inequality is strict (see examples in our simulations).

To make the further developments more concrete, we now present two classical examples: logistic regression and Poisson regression.

\paragraph{Logistic regression.} The special case of conditional family is logistic regression, where $y \in \{0,1\}$,
$a(t) = \log ( 1 + e^{-t})$ and $a'(t) = \sigma(t) = \frac{1}{1+e^{-t}}$ is the sigmoid function and the probability mass function is given by $p(y|\eta) = \exp(\eta y -\log(1+e^{\eta}))$.

\paragraph{Poisson regression.} One more special case is Poisson regression with $y \in \mathbb{N}$, $a(t) = \exp(t)$ and the response variable $y$ has a Poisson distribution.  The probability mass function is given by $p(y|\eta) = \exp(\eta y - e^\eta - \log(y!))$. Poisson regression may be appropriate when the dependent variable is a count, for example in genomics,
network packet analysis, crime rate analysis, fluorescence microscopy, etc.~\citep{hilbe2011negative}.

\subsection{AVERAGING PREDICTIONS}
Recall from \mysec{sgd} that $\pi_\gamma$ is the stationary distribution of $\theta$. 
Taking expectation of both parts of Eq. (\ref{eq:sgd}), we get, by using the fact that $\pi_\gamma$ is the limiting distribution
of $\theta_n$ and $\theta_{n-1}$:
\BEAS
& & \E_{\pi_\gamma(\theta_n)}\theta_n \\
&  =  &  \E_{\pi_\gamma(\theta_{n-1})}\theta_{n-1} -\gamma \E_{\pi_\gamma(\theta_{n-1})}\E_{p(x_n,y_n)}f_n'(\theta_{n-1}),
\EEAS
which leads to $\E_{\pi_\gamma(\theta)} \E_{p(x_n,y_n)} \nabla f_n(\theta)=0$, that is, now removing the dependence on $n$ (data $(x,y)$ are i.i.d.):
$$\E_{\pi_\gamma(\theta)}\E_{p(x,y)}\Big[-y\Phi(x) + a'\big(\Phi(x)^\top\theta\big)\Phi(x) \Big] = 0 ,$$
which finally leads to
\begin{equation}
\E_{p(x)}\Big[\E_{\pi_\gamma(\theta)}a'\big(\Phi(x)^\top\theta\big) - \mu_{**}(x) \Big]\Phi(x) = 0.
\label{limitgrad}
\end{equation}
This is the core equation our method relies on. It does not imply that $b(x) = \E_{\pi_\gamma(\theta)}a'\big(\Phi(x)^\top\theta\big) - \mu_{**}(x)$ is uniformly equal to zero (which we want), but only that 
$\E_{p(x)} \Phi(x) b(x) = 0$, i.e., $b(x)$ is uncorrelated with $\Phi(x)$.

If the feature vector $\Phi(x)$ is ``large enough'' then this 
 is equivalent to $b=0$.\footnote{Let $\Phi(x) = (\phi_1(x),\dots, \phi_n(x))^\top$ be an orthogonal basis and $b(x) = \sum_{i=1}^n c_i\phi_i(x) + \varepsilon(x)$, where $\varepsilon(x)$ is small if the basis is big enough. Then $\mathbb{E}_{p(x)}\Phi(x)b(x) = 0 \Leftrightarrow \mathbb{E}\phi_i(x)\big[\sum_{i=1}^n c_i\phi_i(x) + \varepsilon(x)\big] = 0$ for every $i$, and due to the orthogonality of the basis and the smallness of $\varepsilon(x)$: $c_i\cdot\mathbb{E}_{p(x)}\phi^2(x) \approx 0$ and hence $c_i \approx 0$ and thus $b(x)\approx 0$.} 

   For example, when $\Phi(x)$ is composed of an orthonormal basis of the space of integrable functions (like for kernels in \mysec{kernels}), then this is exactly true. 
Thus, in this situation,
\BEQ
\label{eq:consistency}\mu_{\ast \ast}(x) = \E_{\pi_\gamma(\theta)}a'\big(\Phi(x)^\top\theta\big), 
\EEQ
and averaging predictions $a'\big(\Phi(x)^\top\theta_n\big)$,  along the path $(\theta_n)$ of the Markov chain should exactly converge to the optimal prediction.

This exact convergence is weaker (requires high-dimensional fatures) than for the unconditional family in \mysec{uncond} but it can still bring surprising benefits even when $\Phi$ is not large enough, as we present in \mysec{finited} and \mysec{kernels}.

\subsection{TWO TYPES OF AVERAGING}
\label{sec:avetwotype}
Now we can introduce two possible ways to estimate the prediction function $\mu(x)$.

\paragraph{Averaging estimators.}

The first one is the usual way:   we first estimate parameter $\theta$, using Ruppert-Polyak averaging \citep{polyak1992acceleration}: $\bar{\theta}_n = \frac{1}{n+1}\sum_{i=0}^n\theta_i$ and then we denote 
$$\bar{\mu}_n(x) = a'(  \Phi(x)^\top \bar{\theta}_n )
= a' \Big(
 \Phi(x)^\top\textstyle  \frac{1}{n+1} \sum_{i=0}^n\theta_i
\Big) $$ the corresponding prediction. As discussed in \mysec{sgd} it converges to $\bar{\mu}_\gamma: x \mapsto 
a'(  \Phi(x)^\top \bar{\theta}_\gamma )$, which is \emph{not} equal to in general to $a'(  \Phi(x)^\top \theta_\ast )$, where $\theta_\ast$ is the optimal parameter in $\rb^d$. Since, as presented at the end of \mysec{sgd}, $\bar{\theta}_\gamma - \theta_\ast$ is of order $O(\gamma)$, $F(\bar{\theta}_\gamma) - F(\theta_\ast)$ is of order $O(\gamma^2)$ (because
$\nabla F (\theta_\ast) = 0$), and thus an error of $O(\gamma^2)$ is added to the usual convergence rates in~$O(1/n)$. 

Note  that we are limited here to prediction functions which corresponds to \emph{linear functions} in $\Phi(x)$ in the natural parameterization, and thus $ F(\theta_\ast) \geqslant \mathcal{G}(\mu_{\ast \ast})$, and the inequality is often strict.
 
 \paragraph{Averaging predictions.}
 We propose a new estimator
 $$
 \bar{\bar{\mu}}_n(x) = \frac{1}{n+1} \sum_{i=0}^n a'(\theta_i^\top \Phi(x) ).
 $$
 In general  $  \mathcal{G} (\bar{\bar{\mu}}_n) -\mathcal{G}(\mu_{\ast\ast})$ does not converge to zero either (unless the feature vector $\Phi$ is large enough and \eq{consistency} is satisfied). Thus, on top of the usual convergence in $O(1/n)$ with respect to the number of iterations, we have an extra term that depends only on $\gamma$, which we will study in \mysec{finited} and \mysec{kernels}.

We denote by   $\bar{\bar{\mu}}_\gamma(x)$ the limit when $n \to \infty$, that is, using properties of converging Markov chains,
$\bar{\bar{\mu}}_\gamma(x) = \E_{\pi_\gamma(\theta)}a'\big(\Phi(x)^\top\theta\big)$.

Rewriting   Eq. (\ref{limitgrad}) using our new notations, we get:
 $$
 \E \big[
 ( \mu_{\ast\ast}(x) - \bar{\bar{\mu}}_\gamma(x))
 \Phi(x_n) \big] = 0.
 $$
 When $\Phi : \rb \to \rb^d$ is high-dimensional, this leads to $\mu_{\ast\ast} = \bar{\bar{\mu}}_\gamma$ and in contrast to $\bar{\mu}_\gamma$, averaging predictions potentially converge to the optimal prediction.
 
 \paragraph{Computational complexity.}
Usual averaging of estimators~\citep{polyak1992acceleration} to compute $\bar{\mu}_n(x) = a'(  \Phi(x)^\top \bar{\theta}_n ) $ is simple to implement as we can simply update the average~$\bar{\theta}_n$ with essentially no extra cost on top the complexity $O(nd)$ of the SGD recursion. Given the number $n$ of training data points and the number $m$ of testing data points, the overall complexity is $O( d(n+m))$.

Averaging prediction functions is more challenging as we have to store all iterates $\theta_i$, $i\!=\!1,\dots,n$, and for each testing point $x$, compute  $
 \bar{\bar{\mu}}_n(x) = \frac{1}{n+1} \sum_{i=0}^n a'(\theta_i^\top \Phi(x) ).
 $
 Thus the overall complexity is $O( dn+mnd )$, which could be too costly with many test points (i.e., $m$ large).
 
 There are several ways to alleviate this extra cost: (a) using sketching techniques~\citep{woodruff2014sketching}, (b) using summary statistics like done in applications of MCMC~\citep{gilks1995markov}, or (c) leveraging the fact that all iterates $\theta_i$ will end up being close to $\bar \theta_\gamma$ and use a Taylor expansion of $a'\big(\theta^\top \Phi(x)\big) $ around~$\bar \theta_\gamma$. This expansion is equal to:
$$ a'\big(\Phi(x)^\top \overline{\theta}_\gamma\big) + (\theta-\overline{\theta}_\gamma)^\top  \Phi(x) \cdot a''\big(\Phi(x)^\top \overline{\theta}_\gamma\big)+$$

\vspace*{-1cm}

$$+\frac 12\big( (\theta-\overline{\theta}_\gamma  )^\top \Phi(x) \big)^{2}\cdot a'''\big(\Phi(x)^\top \overline{\theta}_\gamma\big) + O\big(\Vert \theta-\overline{\theta}_\gamma\Vert^3\big). $$
Taking expectation in both sides above leads to:
$$\bar{\bar{\mu}}_\gamma(x) \approx \bar{\mu}_\gamma(x) + \frac{1}{2}\Phi(x)^\top \mbox{cov\,} (\theta)\cdot \Phi(x)\cdot a'''\big(\overline{\theta}_\gamma^\top\Phi(x)\big),$$ 
where $ \mbox{cov\,} (\theta)$ is the covariance matrix of $\theta$ under $\pi_\gamma$. This
  provides a simple correction to $\bar\mu_\gamma$, and leads to an approximation of $\bar{\bar{\mu}}_n(x)$ as 
$$\bar{\mu}_n(x) + \frac 12\ \Phi(x)^\top \mbox{cov}_n(\theta)\ \Phi(x) \cdot a'''\big(\overline{\theta}_n^\top\Phi(x)\big), $$
where $ \mbox{cov}_n(\theta)$ is the empirical covariance matrix of the iterates $(\theta_i)$.

The computational complexity now becomes $O(nd^2 + m d^2)$, which is an improvement when the number of testing points $m$ is large. In all of our experiments, we used this approximation.

\section{FINITE-DIMENSIONAL MODELS}
\label{sec:finited}

In this section we study the behavior of  $ \bar{\bar{A}}(\gamma) =  \mathcal{G} (\bar{\bar{\mu}}_\gamma) -\mathcal{G}(\mu_{\ast})$ for finite-dimensional models, for which it is usually not equal to zero. We know that our estimators $\bar{\bar{\mu}}_n$ will converge to 
$\bar{\bar{\mu}}_\gamma$, and our goal is to compare it to $  \bar{A}(\gamma) = \mathcal{G} ({\bar{\mu}}_\gamma) 
-\mathcal{G}(\mu_{\ast}) = F(\bar{\theta}_\gamma) - F(\theta_\ast)$ which is what averaging estimators tends to. We also consider for completeness the non-averaged performance  $   {A}(\gamma)  = \E_{\pi_\gamma(\theta)} \big[F( {\theta}) - F(\theta_\ast)
\big]$.

Note that we must have ${A}(\gamma)  $ and  $  \bar{A}(\gamma) $ non-negative, because we compare the negative log-likelihood performances to the one of of the best linear prediction (in the natural parameter), while $ \bar{\bar{A}}(\gamma)$ could potentially be negative (it will in certain situations), because the the corresponding natural parameters may not be linear in $\Phi(x)$.

We consider the same standard assumptions as~\citet{dieuleveut2017bridging}, namely smoothness of the negative log-likelihoods $f_n(\theta)$ and strong convexity of the expected negative log-likelihood $F(\theta)$. We first recall the results from~\citet{dieuleveut2017bridging}.  See detailed explicit formulas in the supplementary material.

\subsection{EARLIER WORK}

\paragraph{Without averaging.}
We have that $A(\gamma) = \gamma B + O(\gamma^{3/2})$, that is $\gamma$ is \emph{linear} in $\gamma$, with $B$ non-negative.

\paragraph{Averaging estimators.} We have that $\bar{A}(\gamma) = \gamma^2 \bar{B} + O(\gamma^{5/2})$, that is $\bar{A}$ is \emph{quadratic} in $\gamma$, with $\bar B$ non-negative. Averaging does provably bring some benefits because the order in $\gamma$ is higher (we assume $\gamma$ small).

\subsection{AVERAGING PREDICTIONS}

We are now ready to analyze the behavior of our new framework of averaging predictions. The following result is shown in the supplementary material.

\begin{proposition}
Under the assumptions on the negative loglikelihoods $f_n$ of each observation from~\citet{dieuleveut2017bridging}:
\begin{itemize}
\item In the case of well-specified data, that is, there exists $\theta_\ast$ such that for all $(x,y)$, $p(y|x) = q(y|x,\theta_\ast)$, then $\bar{\bar{A}} \sim \gamma^2\bar{\bar{B}}^{\rm well}$, where $\bar{\bar{B}}^{\rm well}$ is a positive constant.
\item In the general case of potentially mis-specified data, $\bar{\bar{A}} = \gamma\bar{\bar{B}}^{\rm mis} + O(\gamma^2)$, where $\bar{\bar{B}}^{\rm mis}$ is constant which may be positive or negative.
\end{itemize}
\end{proposition}

Note, that in contrast to averaging parameters, the constant $\bar{\bar{B}}^{\rm mis}$ can be negative. It means, that we obtain the estimator better than the optimal linear estimator, which is the limit of capacity for averaging parameters. In our simulations, we show examples for which $\bar{\bar{B}}^{\rm mis}$ is positive, and examples for which it is negative. Thus, in general, for low-dimensional models, averaging predictions can be worse or better than averaging parameters. However, as we show in the next section, for infinite dimensional models, we always get convergence.

\section{INFINITE-DIMENSIONAL MODELS}
\label{sec:kernels}

Recall, that we have the following objective function to minimize:
\begin{equation}
F(\theta) = \E_{x,y}\Big[ -y \cdot \eta_\theta(x) + a\big(\eta_\theta(x)\big)  \Big],
\end{equation}
where till this moment we consider unknown functions $\eta_\theta(x)$ which were linear in $\Phi(x)$ with $\Phi(x) \in \rb^d$, leading to a complexity in $O(dn)$. 

We now consider infinite-dimensional features, by considering that $\Phi(x) \in \mathcal{H}$, where $\mathcal{H}$ is a Hilbert space. Note that this corresponds to modeling the function $\eta_\theta$ as a Gaussian process~\citep{rasmussen2004gaussian}.

This is computationally feasible through the usual  ``kernel trick'' \citep{scholkopf2001learning,shawe2004kernel}, where we assume that the kernel function $k(x,y) = \langle \Phi(x), \Phi(y) \rangle$ is easy to compute.  
Indeed, following~\citet{bordes2005fast} and \citet{Die_Bac_2015}, starting from $\theta_0$, each iterate of constant-step-size SGD is of the form
$\theta_n = \sum_{t=1}^n\alpha_t\Phi(x_t)$, and the gradient descent recursion 
$\theta_n = \theta_{n-1} - \gamma [ a' (\eta_{\theta_{n-1}}(x_n) )-y_n ]  \Phi(x_n)$
leads to the following recursion on $\alpha_t$'s:\BEAS
\alpha_n & = &\textstyle  -\gamma \big[a'\Big(\sum_{t=1}^{n-1}\alpha_t \langle \Phi(x_t), \Phi(x_n)  \rangle\big) - y_n  \big]  \\ 
& = & \textstyle  -\gamma \big[a'\big(\sum_{t=1}^{n-1}\alpha_t k(x_t,x_n) \big) - y_n  \big].
 \EEAS
 This leads to $\eta_{\theta_n}(x)=\langle\Phi(x),\theta_n \rangle$ and $\mu_{\theta_n}(x) = a'\big(\eta_{\theta_n}(x)\big)$
 with 
 $$\eta_{\theta_n}(x) = \sum\limits_{t=1}^n\alpha_t \langle \Phi(x),\Phi(x_t) \rangle =\sum\limits_{t=1}^n\alpha_t k(x,x_t) , $$
and finally we can express $\bar{\bar{\mu}}_n(x)$ in kernel form as:
$$\bar{\bar{\mu}}_n(x)  = \frac{1}{n+1}\sum\limits_{t=0}^na'\Big[\sum\limits_{l=1}^t\alpha_l\cdot k(x,x_l)\Big]. $$
There is also a straightforward estimator for averaging parameters, i.e., $\bar{\mu}_n(x) = a'\Big(\frac {1}{n+1}\sum\limits_{t=0}^n\sum\limits_{l=1}^t\alpha_l k(x,x_l)\Big).$
If we assume that the kernel function is \emph{universal}, that is, $\mathcal{H}$ is dense in the space of squared integrable functions, then it is known that if $\E_x b(x) \Phi(x) = 0$, then $b=0$~\citep{sriperumbudur2008injective}. This implies that we must have $\bar{\bar{\mu}}_\gamma =0$ and thus averaging predictions does always converge to the global optimum (note that in this setting, we must have a well-specified model because we are in a non-parametric setting).

\paragraph{Column sampling.}

Because of the usual high running-time complexity of kernel method in $O(n^2)$, we consider a ``column-sampling approximation''~\citep{williams2001using}. We thus choose a small subset $I = (x_1,\dots, x_m)$ of samples and construct a new finite $m$-dimensional feature map   $\bar{\Phi}(x) = K(I,I)^{-1/2}K(I,x) \in \R^{m }$, where $K(I,I)$ is the $m \times m$ kernel matrix of the $m$ points and 
$K(I,x) $ the vector composed of kernel evaluations $k(x_i,x)$. This allows a running-time complexity in $O(m^2n)$. In theory and practice, $m$ can be chosen small~\citep{bach2013sharp,rudi2017falkon}.

\paragraph{Regularized learning with kernels.}
Although we can use an unregularized recursion with good convergence properties~\citep{Die_Bac_2015}, adding a regularisation by the squared Hilbertian norm is easier to analyze and more stable with limited amounts of data. We thus consider the recursion (in Hilbert space), with $\lambda$ small:
\BEAS
 \theta_n \!\!\!& = &   \!\!\!\theta_{n-1} - \gamma \big[  f_n'(\theta_{n-1}) + \lambda \theta_{n-1} \big]
 \\
 & = &  \!\!\! \theta_{n-1} + \gamma  ( y_n -  a'( \langle\Phi(x_n),  \theta \rangle )   ) \Phi(x_n) - \gamma \lambda\theta_{n-1} .
 \EEAS
 This recursion can also be computed efficiently as above using the kernel trick and column sampling approximations.
 
 In terms of convergence, the best we can hope for is to converge to the minimizer $\theta_{\ast ,\lambda}$ of the regularized expected negative log-likelihood $F(\theta) + \frac{\lambda}{2} \| \theta\|^2$ (which we assume to exist). When $\lambda$ tends to zero, then 
 $\theta_{\ast ,\lambda}$  converges to $\theta_\ast$. 
 
 Averaging \emph{parameters} will tend to a limit $\bar{\theta}_{\gamma,\lambda}$ which is $O(\gamma)$-close to  $\theta_{\ast ,\lambda}$, thus leading to predictions which deviate from the optimal predictions for two reasons: because of regularization and because of the constant step-size. Since $\lambda$ should decrease as we get more data, the first effect will vanish, while the second will not.

 When averaging \emph{predictions}, the two effects will vanish as $\lambda$ tends to zero. Indeed, by taking limits of the gradient equation, and denoting by $\bar{\bar{\mu}}_{\gamma,\lambda}$  the limit of $\bar{\bar{{\mu}}}_n$, we  have
\begin{equation}
 \E \big[
 ( \mu_{\ast\ast}(x) - \bar{\bar{\mu}}_{\gamma,\lambda}(x))
 \Phi(x) \big] = \lambda \bar{\theta}_{\gamma, \lambda}.
 \label{RegKer}
\end{equation}
Given that $\bar{\theta}_{\gamma, \lambda}$ is $O(\gamma)$-away from $\theta_\ast$, if we assume that $\theta_\ast$ corresponds to a sufficiently regular\footnote{While our reasoning  is informal here, it can be made more precise by considering so-called ``source conditions'' commonly used in the analysis of kernel methods~\citep{caponnetto2007optimal}, but this is out of the scope of this paper.} element of the Hilbert space $\mathcal{H}$, then  the $L_2$-norm of the deviation satisfies $ \| \mu_{\ast\ast}(x) - \bar{\bar{\mu}}_{\gamma,\lambda} \|  = O(\lambda)$  and thus as the regularization parameter $\lambda$ tends to zero, our predictions tend to the optimal one.

\section{EXPERIMENTS}
\label{sec:exp}
In this section, we compare the two types of averaging (estimators and predictions) on a variety of problems, both on synthetic data and on standard benchmarks. When averaging predictions, we always consider the Taylor expansion approximation presented at the end of \mysec{avetwotype}.

\subsection{SYNTHETIC DATA} 

\paragraph{Finite-dimensional models.}
 we consider the following logistic regression model: 
$$q(y|x,\theta) = \exp\big(y\cdot \eta_\theta(x) - a(\eta_\theta(x))\big), $$
where we consider a linear model $\eta_\theta(x) = \theta^\top x$ in $x$ (i.e., $\Phi(x) = x$), the link function $a(t) = \log(1+e^t)$ and $a'(t) = \sigma(t)$ is the sigmoid function. 
Let $x$ be distributed as a standard normal random variable in  dimension $d=2$, $y\in \{0,1\}$ and $\P(y=1|x) = \mu_{\ast\ast}(x) = \sigma\big(\eta_{\ast\ast}(x)\big)$, where we consider two different settings:
\begin{itemize}
\item Model 1: $\eta_{\ast\ast}(x) = \sin x_1 + \sin x_2$,
\item Model 2: $\eta_{\ast\ast}(x) = x_1^3+x_2^3 $.
\end{itemize}
The global minimum $\FF_{\ast\ast}$ of the corresponding optimization problem can be found as
$$\FF_{\ast\ast}    = \E_{p(x)}\big[-\mu_{\ast\ast}(x)\cdot \eta_{\ast\ast}(x) + a(\eta_{\ast\ast}(x))\big]. $$
We also introduce the performance measure $\FF(\eta)$ 
\begin{equation}
\FF(\eta) = \E_{p(x)}\big[-\mu_{\ast\ast}(x)\cdot \eta(x) + a(\eta(x)) \big],
\label{Fofeta}
\end{equation}
which can be evaluated directly in the case of synthetic data. Note that in our situation, the model is misspecified because $\eta_{\ast\ast}(x)$ is not linear in $\Phi(x) = x$, and thus, $\inf_\theta F(\theta) > \FF_{\ast\ast} $, and thus our performance measures $\FF(\mu_n) - \FF_{\ast\ast}  $ for various estimators $\mu_n$ will not converge to zero.

\begin{figure}
\includegraphics[width=0.5\textwidth]{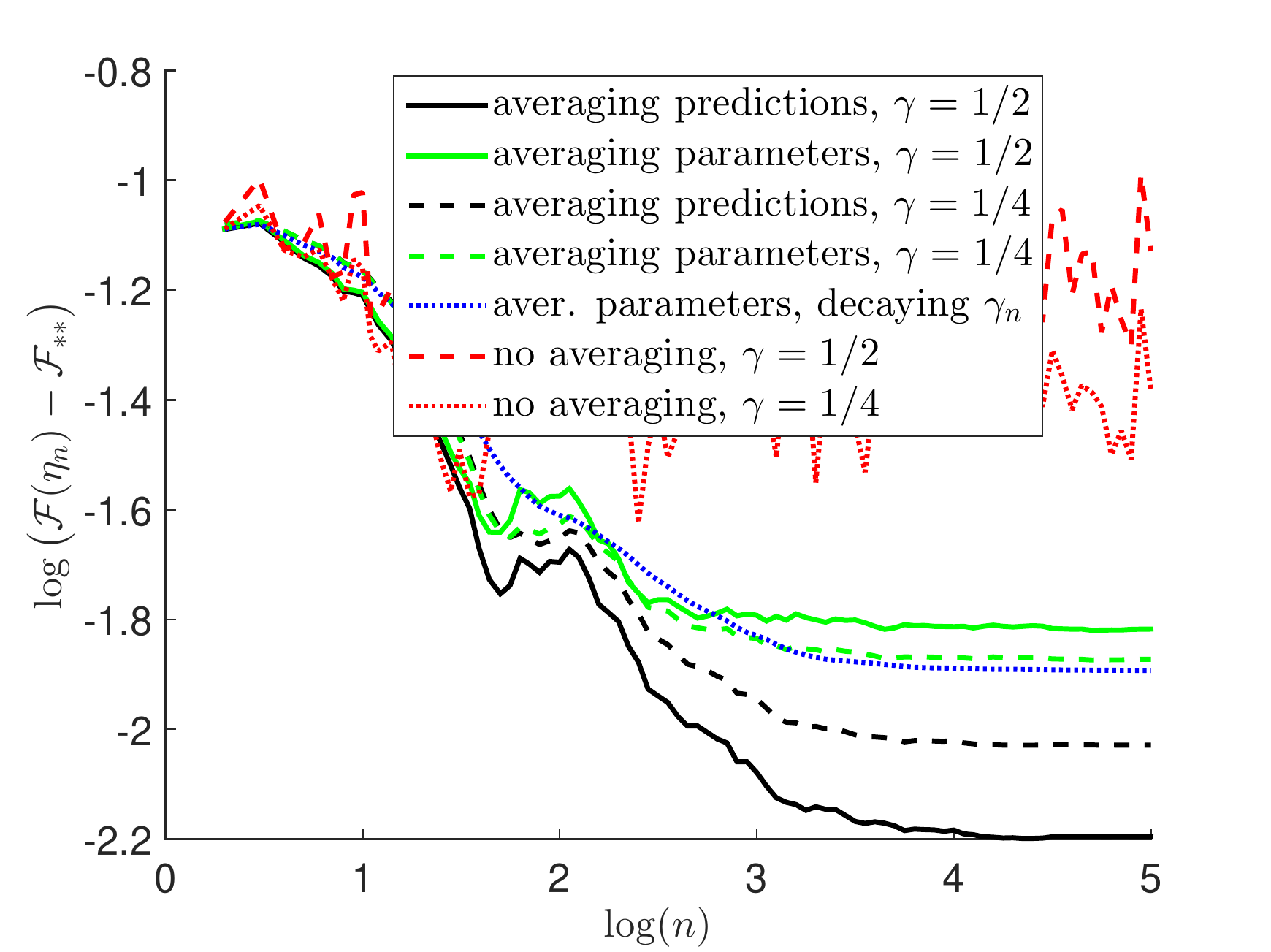}

\vspace*{-.415cm}

\caption{Synthetic data for linear model $\eta_\theta(x) = \theta^\top x $ and $\eta_{\ast\ast}(x) = \sin x_1 + \sin x_2$. Excess prediction performance vs. number of iterations (both in log-scale).}
\label{Graph01}
\end{figure}

\begin{figure}
\includegraphics[width=0.5\textwidth]{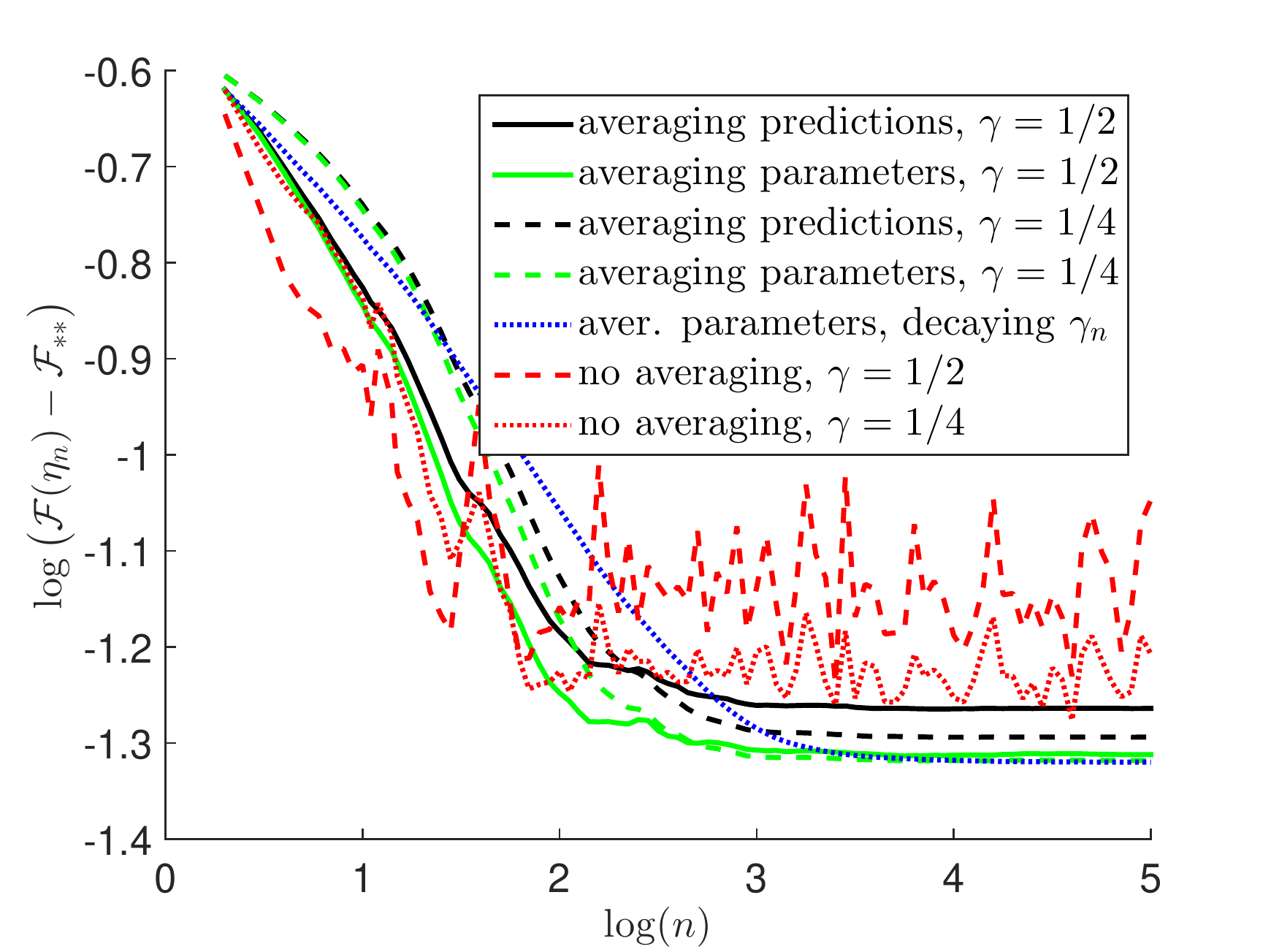}

\vspace*{-.415cm}

\caption{Synthetic data for linear model $\eta_\theta(x) = \theta^\top x $ and $\eta_{\ast\ast}(x) = x_1^3+x_2^3$.  Excess prediction performance vs. number of iterations (both in log-scale). }
\label{Graph02}
\end{figure}

The results of averaging 10 replications are shown in Fig.~\ref{Graph01} and Fig.~\ref{Graph02}. 
We first observed that constant step-size SGD without averaging leads to a bad performance.

Moreover, we can see, that in the first case (Fig.~\ref{Graph01}) averaging predictions beats averaging parameters, and moreover beats the best linear model: if we use the best linear error $\FF_{*}$ instead of $\FF_{**}$, at some moment $\FF(\eta_n)-\FF_{\ast}$ becomes negative. However in the second case (Fig.~\ref{Graph02}), averaging predictions is not superior to averaging parameters. Moreover, by looking at the final differences between performances with different values of $\gamma$, we can see the dependency of the final performance in $\gamma$ for averaging predictions, instead of $\gamma^2$ for averaging parameters (as suggested by our theoretical results in \mysec{finited}). In particular in Fig.~\ref{Graph01}, we can observe the surprising behavior of a larger step-size leading to a better performance (note that we cannot increase too much otherwise the algorithm would diverge).

\begin{figure}
\includegraphics[width=0.47\textwidth]{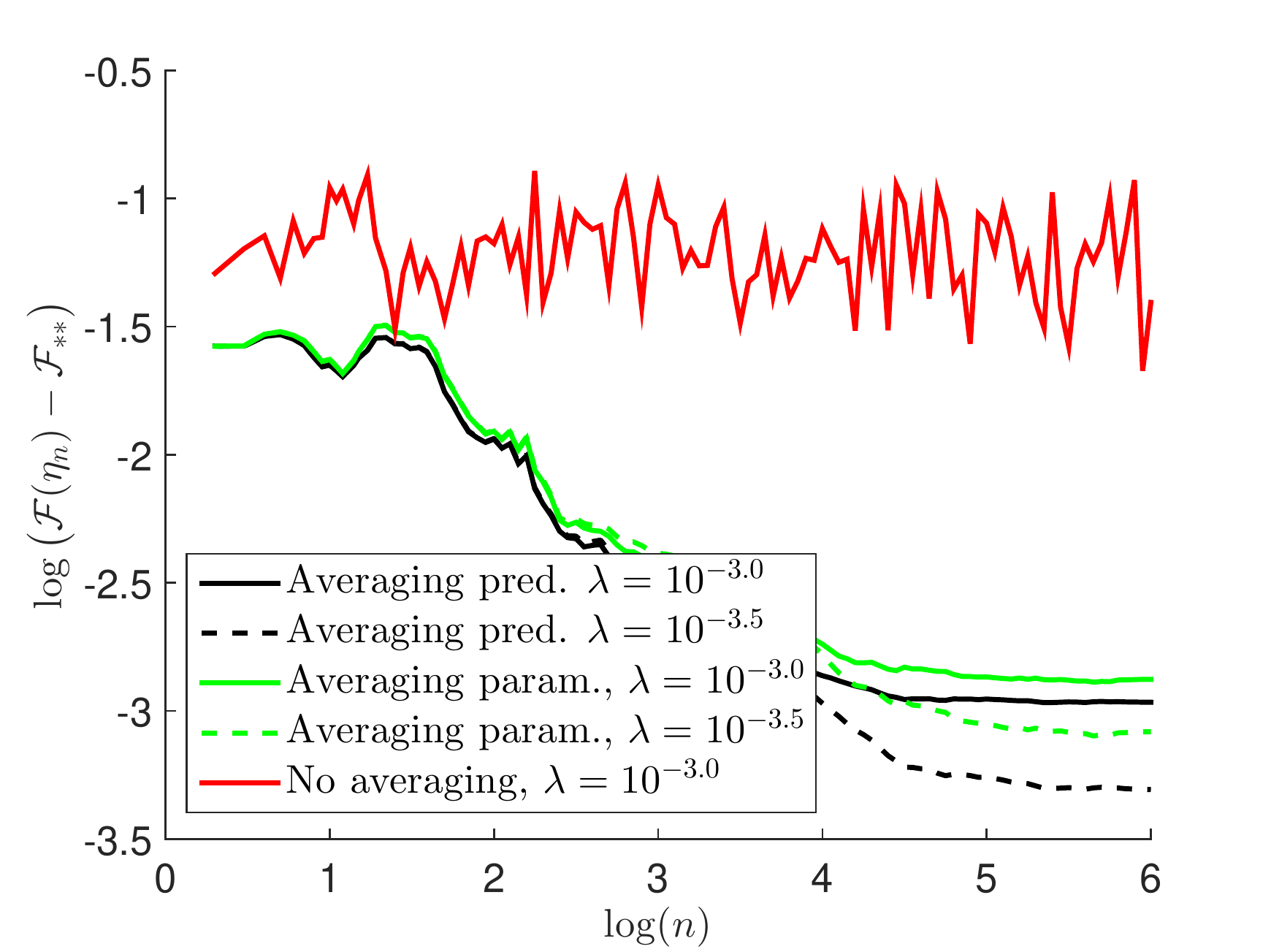}

\vspace*{-.415cm}

\caption{Synthetic data for Laplacian kernel for $\eta_{\ast\ast}(x) = \frac{5}{5+x^\top x}$. Excess prediction performance vs. number of iterations (both in log-scale). }
\label{Graph03}
\end{figure}

\paragraph{Infinite-dimensional models} Here we consider the kernel setup described in \mysec{kernels}. We consider Laplacian kernels $k(s,t) = \exp\big(\frac{\Vert s-t\Vert_1}{\sigma}\big)$ with $\sigma = 50$, dimension $d=5$ and generating log odds ratio $\eta_{\ast\ast}(x) = \frac{5}{5+x^\top x}$. We also use a squared norm regularization with several values of $\lambda$ and column sampling with $m=100$ points. We use the exact value of $\FF_{\ast \ast}$ which we can compute directly for synthetic data. The results are shown in Fig.~\ref{Graph03}, where averaging predictions leads to a better performance than averaging estimators.

\subsection{REAL DATA}

Note, that in the case of real data, one does not have access to unknown $\mu_{\ast\ast}(x)$ and computing the performance measure in Eq.~(\ref{Fofeta}) is inapplicable. Instead of it we use its sampled version on held out data:
$$\hat{\FF}(\eta) = -\sum\limits_{i:y_i=1}\log\big(\mu(x_i)\big) - \sum\limits_{j:y_j=0}\log\big(1-\mu(x_i)\big). $$
We use two  datasets, with $d$ not too large, and $n$ large, from \citep{Lichman2013}: the ``MiniBooNE particle identification'' dataset ($d=50$, $n=130\ 064$), the ``Covertype'' dataset ($d=54$, $n=581\ 012$).

\begin{figure}
\includegraphics[width=0.5\textwidth]{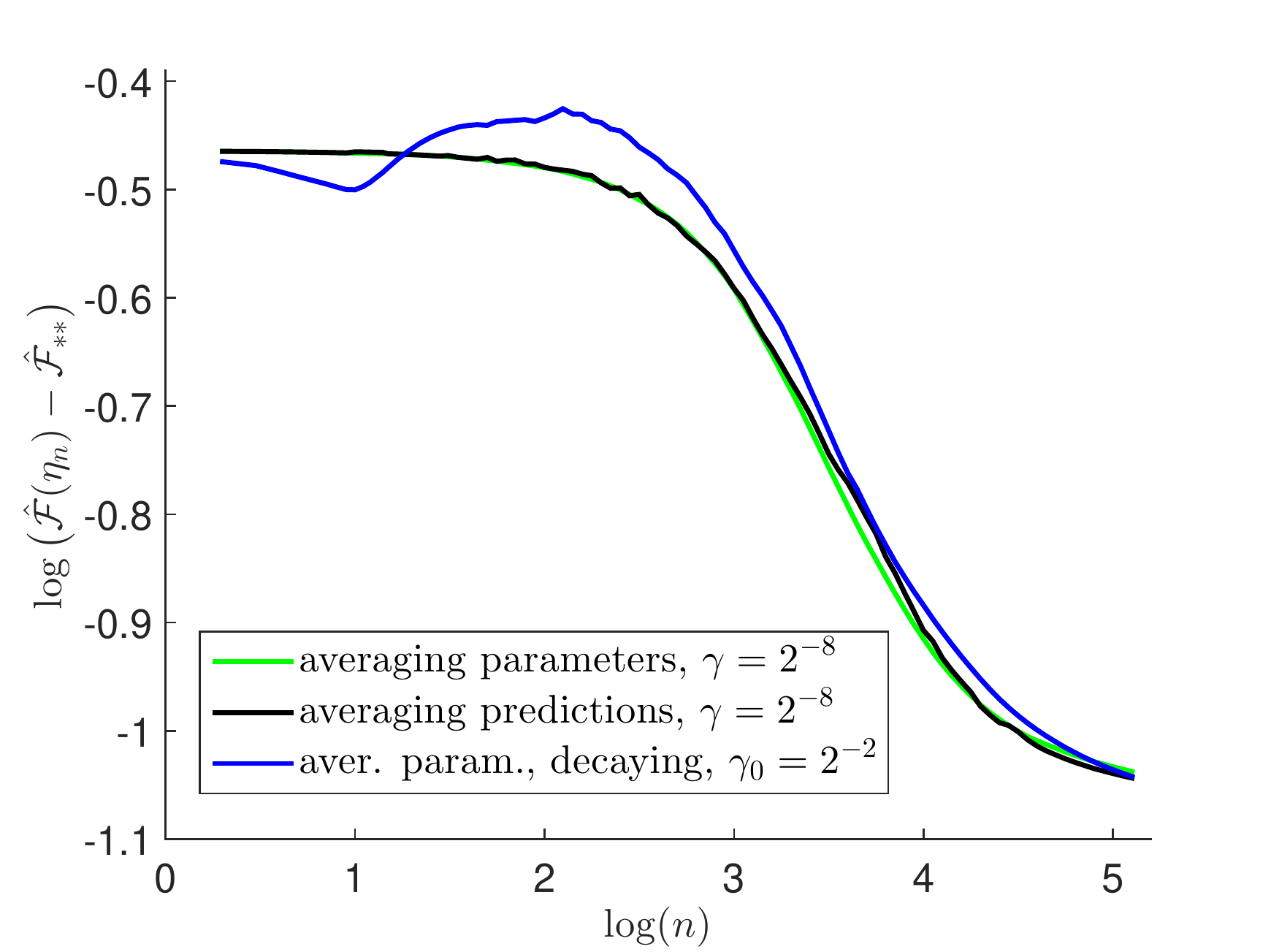}

\vspace*{-.415cm}

\caption{MiniBooNE dataset, dimension $d=50$, linear model. Excess prediction performance vs. number of iterations (both in log-scale).}
\label{Graph04}
\end{figure}

\begin{figure}
\includegraphics[width=0.5\textwidth]{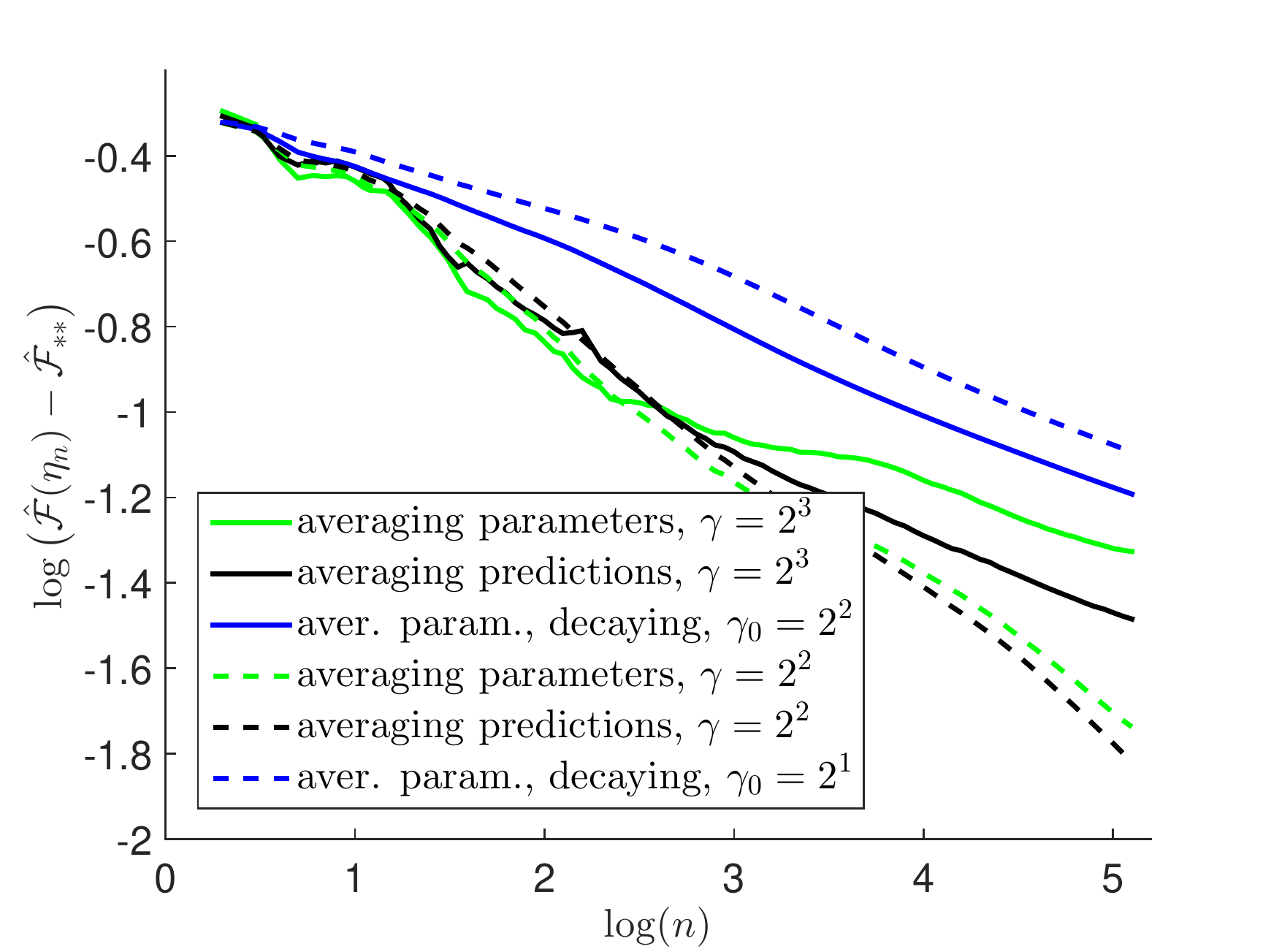}

\vspace*{-.415cm}

\caption{MiniBooNE dataset, dimension $d=50$, kernel approach, column sampling $m=200$. Excess prediction performance vs. number of iterations (both in log-scale).}
\label{Graph05}
\end{figure}

\begin{figure}
\includegraphics[width=0.5\textwidth]{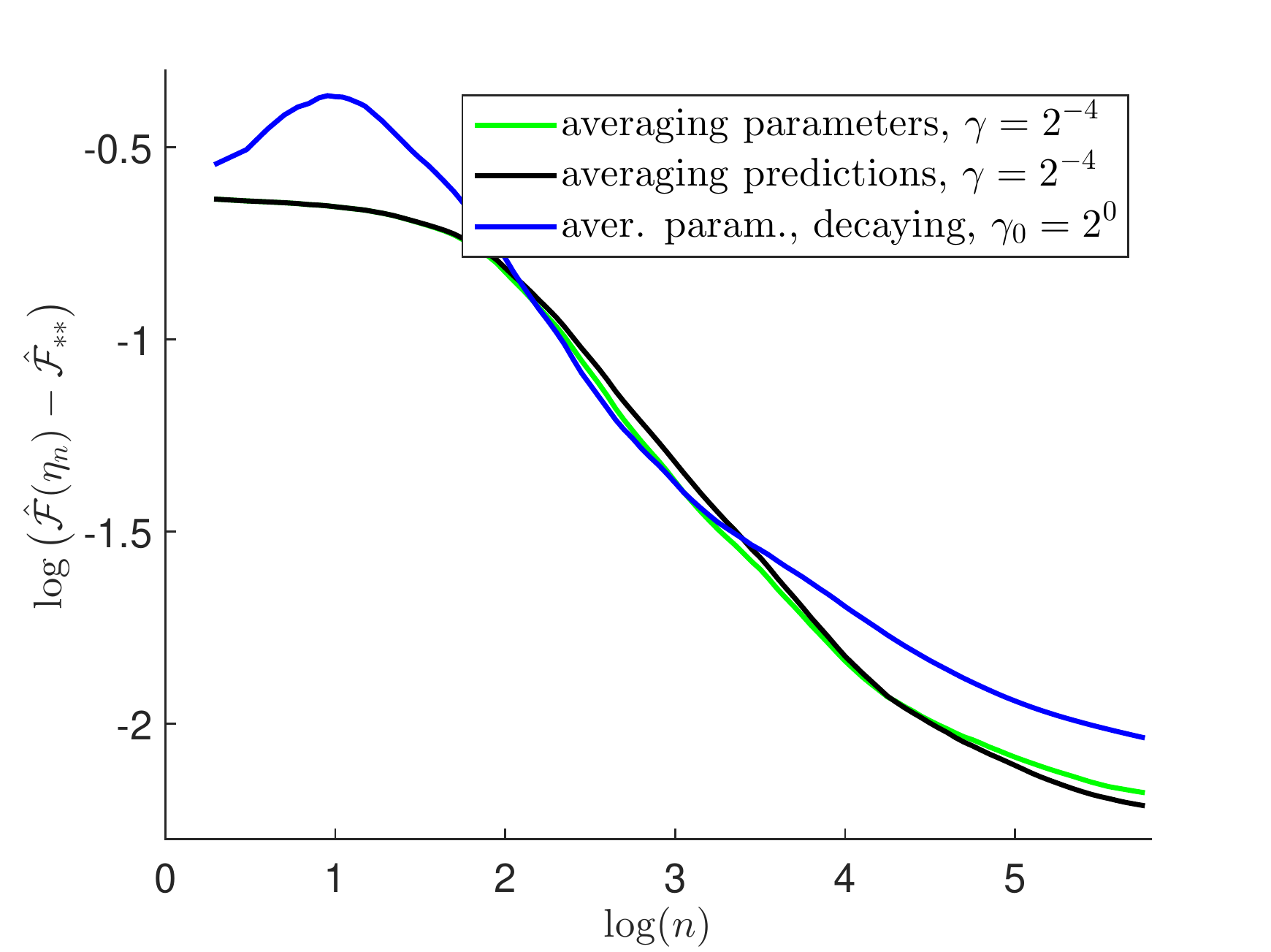}

\vspace*{-.415cm}

\caption{CoverType dataset, dimension $d=54$, linear model. Excess prediction performance vs. number of iterations (both in log-scale).}
\label{Graph06}
\end{figure}

\begin{figure}
\includegraphics[width=0.5\textwidth]{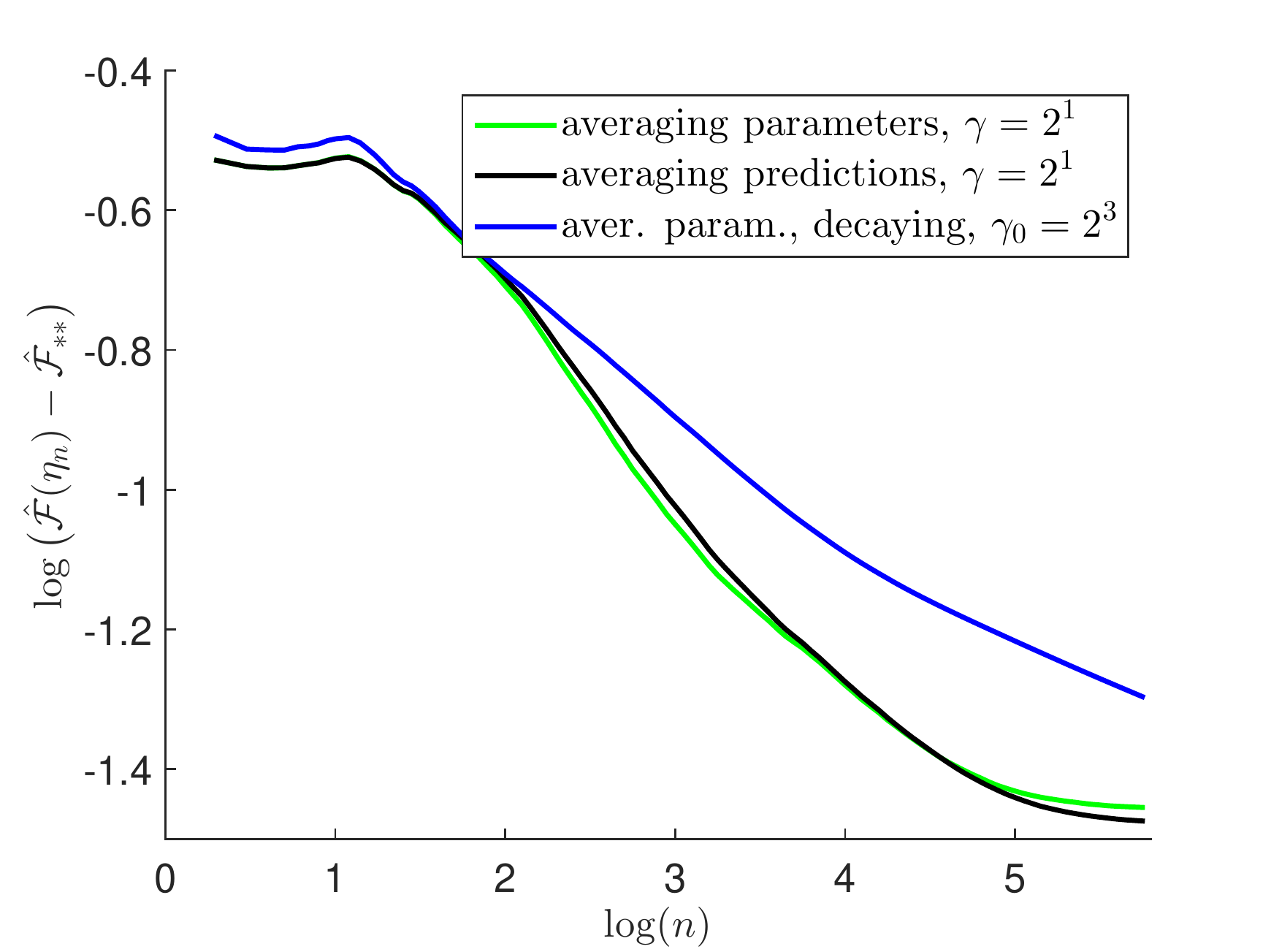}
\caption{CoverType dataset, dimension $d=54$, kernel approach, column sampling $m=200$. Excess prediction performance vs. number of iterations (both in log-scale).}
\label{Graph07}
\end{figure}

We use two different approaches for each of them: a linear model $\eta_\theta(x)=\theta^\top x$ and a kernel approach with Laplacian kernel $k(s,t) = \exp\big(\frac{\Vert s-t\Vert_1}{\sigma}\big)$, where $\sigma =  {d}$. The results are shown in  Figures \ref{Graph04} to \ref{Graph07}.
 Note, that for linear models we use $\hat{\FF}_*$--the estimator of the best performance among linear models (learned on the test set, and hence not reachable from learning on the training data), and for kernels we use $\hat{\FF}_{\ast\ast}$ (same definition as $\hat{\FF}_*$ but with the kernelized model), that is why graphs are not comparable (but, as shown below, the value of $\hat{\FF}_{\ast\ast}$ is lower than the value of $\hat{\FF}_*$ because using kernels correspond to a larger feature space). 
 
For the ``MiniBooNE particle identification'' dataset $\hat{\FF}_{\ast} = 0.35$ and $\hat{\FF}_{\ast\ast} = 0.21; $ for the``Covertype'' dataset $\hat{\FF}_{\ast} = 0.46$ and $\hat{\FF}_{\ast\ast} = 0.39$.  We can see from the four plots  that, especially in the kernel setting, averaging predictions also shows better performance than averaging parameters.

\section{CONCLUSION}
In this paper, we have explored how  averaging procedures in stochastic gradient descent, which are crucial for fast convergence, could be improved by looking at the specifics of probabilistic modeling. Namely, averaging in the moment parameterization can have better properties than averaging in the natural parameterization.

While we have   provided some theoretical arguments (asymptotic expansion in the finite-dimensional case, convergence to optimal predictions in the infinite-dimensional case), a detailed theoretical analysis with explicit convergence rates would provide a better understanding of the benefits of averaging predictions.

\subsection*{Acknowledgements} The research leading to these results has received funding from the European Union's H2020 Framework Programme (H2020-MSCA-ITN-2014) under grant agreement n$^o$ 642685 MacSeNet, and from 
 the European Research Council (grant SEQUOIA 724063).

\bibliography{bib}{}
\bibliographystyle{plainnat}

\end{document}